\documentclass[conference]{IEEEtran}
\makeatletter
\let\old@ps@headings\ps@headings
\let\old@ps@IEEEtitlepagestyle\ps@IEEEtitlepagestyle
\def\confheader#1{%
\def\ps@IEEEtitlepagestyle{%
\old@ps@IEEEtitlepagestyle%
\def\@oddhead{\strut\hfill#1\hfill\strut}%
\def\@evenhead{\strut\hfill#1\hfill\strut}%
}%
\ps@headings%
}
\makeatother

\title{NyayaAI: An AI-Powered Legal Assistant Using Multi-Agent Architecture and Retrieval-Augmented Generation}

\IEEEoverridecommandlockouts
\usepackage{cite}
\usepackage{amsmath,amssymb,amsfonts}
\usepackage{algorithmic}
\usepackage{graphicx}
\usepackage{url}
\usepackage{textcomp}
\usepackage{xcolor}
\usepackage{comment}
\usepackage{amsmath}
\usepackage{devanagari}
\usepackage{float}
\usepackage{footnote}
\usepackage{tablefootnote}
\usepackage{dirtytalk}
 \usepackage{algorithm} 
\usepackage{mathtools, nccmath}
\author{

\IEEEauthorblockN{Deepanshu}
\IEEEauthorblockA{\textit{School of Computer Science} \\
\textit{UPES, Dehradun, India} \\
deepanshumiglani0408@gmail.com
}
\and
\IEEEauthorblockN{Divi Saxena}
\IEEEauthorblockA{\textit{School of Computer Science} \\
\textit{UPES, Dehradun, India} \\
divisaxena04@gmail.com
}
\and
\IEEEauthorblockN{Deepali Rana}
\IEEEauthorblockA{\textit{School of Computer Science} \\
\textit{UPES, Dehradun, India} \\
ranadeepali45@gmail.com
}
\and
\IEEEauthorblockN{Ayesha Varshney}
\IEEEauthorblockA{\textit{School of Computer Science} \\
\textit{UPES, Dehradun, India} \\
ayeshavarshney245@gmail.com
}
\and
\IEEEauthorblockN{Sahinur Rahman Laskar}
\IEEEauthorblockA{\textit{School of Computer Science} \\
\textit{UPES, Dehradun, India} \\
sahinurlaskar.nits@gmail.com
}

}

\usepackage[pscoord]{eso-pic}

\begin{document}

\maketitle
\begin{abstract}
Legal information in India remains largely inaccessible due to the complexity of legal language and the sheer volume of legal documentation involved in research and case analysis. This paper presents NyayaAI, an AI-powered legal assistant that automates and simplifies legal workflows for lawyers, law students, and general users. The system combines Large Language Models with a Retrieval-Augmented Generation pipeline grounded in a curated Indian legal knowledge base comprising constitutional provisions, statutes, case laws, and judicial precedents. A multi-agent architecture orchestrated through the Mastra TypeScript framework coordinates a main agent with specialized sub-agents handling legal research, document summarization, case law retrieval, and drafting assistance. A compliance module validates all responses before delivery. Domain classification achieved 70\% precision across test samples, with RAG retrieval precision at 74\% and overall response accuracy at 72\%, demonstrating that structured multi-agent LLM systems can meaningfully improve legal accessibility and workflow efficiency. The code\footnote{https://github.com/B97784/NyayaAI} is made publicly available for the benefit of the research community.

    

\end{abstract}


\section{Introduction}
\label{intro}
Access to legal knowledge in India has historically been constrained by the highly technical nature of legal language, the enormous volume of legal documentation, and the time-intensive demands of legal research. Lawyers spend considerable effort manually reviewing statutes, judgments, and precedents, leaving less time for strategic work and client representation. Law students struggle to navigate complex legal texts without structured guidance, while the general public often cannot afford legal consultation, leaving them unable to assert their rights effectively \cite{surden2019ai}.

The emergence of Large Language Models and Retrieval-Augmented Generation presents an opportunity to address these challenges systematically \cite{lewis2020rag}. While platforms such as Harvey AI and CoCounsel have demonstrated strong LLM-based legal assistance in Western contexts, the Indian legal domain remains significantly underserved by intelligent AI tools \cite{katz2023bar}. NyayaAI addresses this gap by combining multi-agent orchestration, semantic retrieval, and compliance monitoring into a unified legal assistance platform tailored specifically to Indian legal resources and user needs \cite{bommarito2017}.

The key contributions of this work are as follows:
\begin{itemize}
    \item \textbf{Multi-Agent Pipeline}: A Mastra-orchestrated architecture with a main Nyaya agent routing queries to specialized sub-agents for legal research, summarization, case analysis, and drafting.
    \item \textbf{RAG Knowledge Base}: A curated Indian legal vector knowledge base comprising constitutional provisions, statutes, case laws, and judicial precedents enabling grounded, hallucination-reduced responses \cite{lewis2020rag}.
    \item \textbf{Compliance Module}: An independent validation layer ensuring all responses remain within legal, ethical, and jurisdictional boundaries before delivery.
    \item \textbf{Domain Classification}: An automated classifier achieving 70\% precision across constitutional, criminal, civil, family, and corporate law domains, guiding precise agent routing \cite{surden2019ai}.
    \item \textbf{Indian LegalTech Focus}: A scalable platform specifically designed for Indian legal resources, addressing a clear gap in existing LegalTech systems \cite{anthropic2024}.
\end{itemize}

This paper is organized as follows: Section \ref{related} reviews existing legal AI systems. Section \ref{method} describes the system architecture. Section \ref{er} presents evaluation results, and Section \ref{con} concludes with future directions.

\section{Related Works}
\label{related}
\subsection{Keyword-Based Legal Information Systems}
The earliest and most widely used legal information systems were built around keyword-based search engines. Platforms such as Manupatra, SCC Online, and Indian Kanoon digitized large volumes of Indian legal content and made them searchable through basic text matching \cite{bommarito2017}. While these platforms represented a significant step forward from manual book-based research, they remained fundamentally limited in their ability to understand the semantic intent behind a legal query.A lawyer searching for judgments related to a specific legal principle would receive results based solely on the presence of matching keywords rather than contextual relevance, often requiring extensive manual filtering to identify truly useful documents \cite{surden2019ai}.

\subsection{NLP-Based Legal Tools}
The emergence of Natural Language Processing brought a new generation of legal tools capable of performing more intelligent operations on legal text \cite{vaswani2017attention}. Systems began to incorporate document classification, named entity recognition, and basic summarization capabilities, allowing users to interact with legal content in more structured ways. However these NLP-based systems were largely task-specific, designed to perform one function well such as contract review or clause extraction, without offering a unified conversational interface through which users could ask diverse legal questions and receive integrated responses \cite{surden2019ai}.

\subsection{LLM-Based Legal Assistance Platforms}
The introduction of Large Language Model-based legal assistants marked the most significant shift in the existing system landscape. Global platforms such as Harvey AI, CoCounsel, and LexisNexis Protege began leveraging GPT-based models to provide conversational legal research, document drafting assistance, and case summarization capabilities \cite{katz2023bar}. These systems demonstrated that LLMs could effectively interpret complex legal language and generate coherent, contextually appropriate legal responses \cite{lewis2020rag}. However the majority of these platforms were developed exclusively around Western legal systems, particularly those of the United States and the United Kingdom, leaving the Indian legal domain significantly underserved \cite{bommarito2017}.

Within the Indian context specifically, existing legal AI tools remain limited in both number and capability. Most available platforms focus narrowly on document search and retrieval without offering intelligent query understanding, multi-agent task handling, or RAG-grounded response generation \cite{lewis2020rag}. Furthermore existing systems largely operate as monolithic single-model pipelines that handle all query types through a single generalized interface, limiting their ability to handle specialized legal subtasks with the precision that legal professionals require \cite{mastra2024}.

The limitations of existing systems across three key dimensions — namely their lack of semantic understanding in keyword-based platforms, their narrow task specificity in NLP-based tools, and their absence of Indian legal focus in LLM-based platforms — collectively define the problem space that NyayaAI is designed to address \cite{surden2019ai, anthropic2024}.

\section{System Description}
\label{method}
\begin{figure}[htbp]
\centering
\includegraphics[width=\columnwidth]{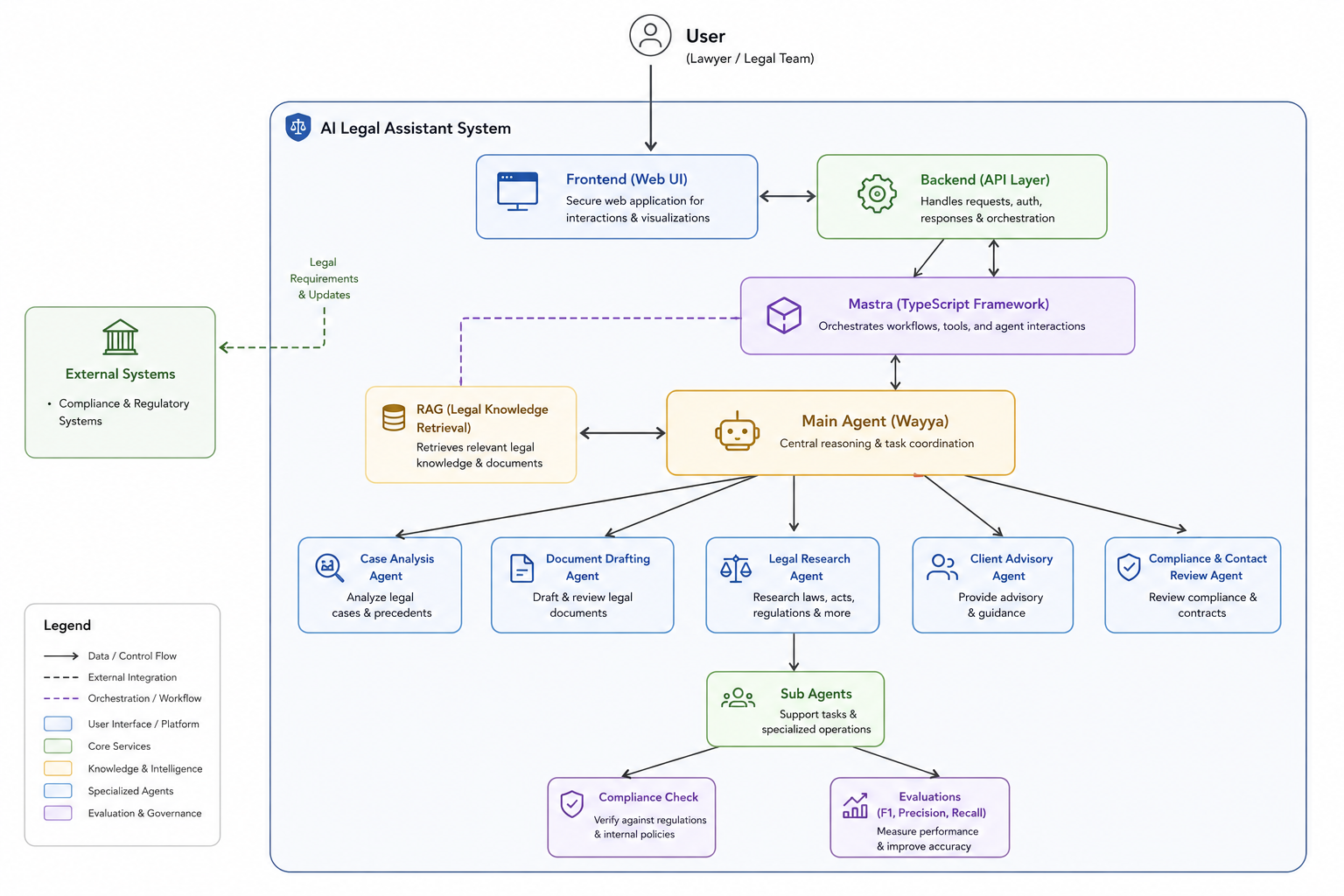}
\caption{NyayaAI System Architecture}
\label{fig:architecture}
\end{figure}

NyayaAI is built as a modular multi-agent pipeline comprising five layers: a Next.js frontend, a Python backend, the Mastra TypeScript orchestration framework, a multi-agent processing layer, and a RAG-powered legal knowledge base. Figure \ref{fig:architecture} illustrates the complete system architecture.

Legal knowledge base preparation involves collecting, preprocessing, and storing curated Indian legal documents as vector embeddings. The backend processing stage manages communication between all system components through well-defined API interfaces. The multi-agent task handling component focuses on intelligent query routing through the Mastra Framework \cite{mastra2024}, delegating specialized legal tasks to dedicated sub-agents. Finally, compliance monitoring and response delivery validates all generated responses before returning them to the user through the Next.js frontend interface.

The Claude SDK was selected as the core language model engine due to its strong capabilities in understanding complex legal language, its support for multi-turn conversational interaction, and its demonstrated reliability in domain-specific reasoning tasks \cite{anthropic2024}. Its efficient integration with the Mastra Framework enables seamless orchestration of the multi-agent pipeline, making it well-suited for legal assistance applications.



\subsection{Technology Stack}
The frontend is developed in Next.js, providing a responsive conversational interface through which users submit legal queries in natural language \cite{mastra2024}. The Python backend manages API communication between all system components and maintains session context for multi-turn interaction. The Mastra TypeScript framework serves as the intelligent orchestration layer, coordinating task delegation across agents and consolidating outputs into unified responses. The Claude SDK powers the core language model engine, selected for its strong legal language comprehension and multi-turn reasoning capabilities \cite{anthropic2024}.

\subsection{Multi-Agent Architecture}
The main orchestrator agent Nyaya receives queries from the Mastra framework and classifies them by legal domain before routing. Simple queries are handled directly by Nyaya, while complex queries are delegated to specialized sub-agents: a legal research agent for statute and case law retrieval, a summarization agent for condensing lengthy legal documents, a case analysis agent for identifying relevant precedents, and a drafting agent for generating preliminary legal document content \cite{mastra2024}. Sub-agent outputs are consolidated by the orchestrator and passed through the compliance module for validation before delivery.

\subsection{RAG Pipeline}
When a sub-agent receives a delegated query, it invokes the RAG module to perform semantic similarity search across the vector knowledge base using FAISS indexing \cite{lewis2020rag}. Retrieved legal documents including constitutional provisions, statutes, and case laws are assembled as contextual input to the language model, ensuring all responses are grounded in verified Indian legal information rather than relying solely on parametric model knowledge \cite{vaswani2017attention}. 

\subsection{Compliance Module}
All agent-generated responses are intercepted by the compliance module before delivery. The module validates responses against defined legal, ethical, and jurisdictional rules, filtering misleading or out-of-scope content and appending appropriate disclaimers where necessary \cite{surden2019ai}.

\section{Results and Analysis}
\label{er}
System performance was evaluated using EVALS metrics across the curated Indian legal knowledge base. As shown in Table \ref{tab:performance_comparison}, the domain classification module achieved 70\% precision, RAG retrieval precision reached 74\%, overall response accuracy was 72\%, and the system F1-score was 0.71 \cite{lewis2020rag, surden2019ai}.

\begin{table}[h]
\centering
\caption{NyayaAI System Evaluation Results}
\label{tab:performance_comparison}
\begin{tabular}{|l|c|}
\hline
\textbf{Metric} & \textbf{Score} \\
\hline
Domain Classification Precision & 70\% \\
RAG Retrieval Precision & 74\% \\
Response Accuracy & 72\% \\
F1-Score & 0.71 \\
\hline
\end{tabular}
\end{table}

Domain-wise analysis revealed Criminal Law (75\%) and Constitutional Law (73\%) achieved the highest classification precision due to well-defined terminology and strong knowledge base representation. Corporate Law recorded the lowest precision (65\%) owing to jurisdictional overlap and limited dataset coverage. Error analysis identified legal jargon complexity (35\%), jurisdictional ambiguity (28\%), context misunderstanding (22\%), and out-of-domain queries (15\%) as primary sources of system limitations, as shown in Table \ref{tab:errors}.

\begin{table}[h]
\centering
\caption{Error Category Distribution}
\label{tab:errors}
\begin{tabular}{|l|c|}
\hline
\textbf{Error Type} & \textbf{Percentage} \\
\hline
Legal Jargon Complexity & 35\% \\
Jurisdictional Ambiguity & 28\% \\
Context Misunderstanding & 22\% \\
Out-of-Domain Queries & 15\% \\
\hline
\end{tabular}
\end{table}

\section{Conclusion and Future Scope}
\label{con}
This paper presented NyayaAI, a multi-agent AI-powered legal assistant that combines the Mastra TypeScript framework, Retrieval-Augmented Generation, and a curated Indian legal knowledge base to deliver accurate, grounded, and compliance-validated legal assistance. The system achieved 70\% domain classification precision and 74\% RAG retrieval precision, demonstrating that structured multi-agent LLM architectures can meaningfully improve legal accessibility and research efficiency in the Indian context \cite{lewis2020rag, surden2019ai}.

Future development directions include integration with national legal databases such as the Supreme Court judgment repository and India Code portal for broader knowledge coverage, introduction of multilingual support for Hindi and other regional Indian languages to expand accessibility, implementation of real-time legal update pipelines to keep the knowledge base current with evolving legislation and judicial decisions, development of predictive analytics for legal outcomes based on historical case law patterns, and support for user-uploaded document analysis enabling AI-assisted review of case-specific legal materials \cite{bommarito2017, mastra2024}.





\bibliographystyle{IEEEtran}
\bibliography{mybib}

@article{surden2019ai,
  author  = {Surden, Harry},
  title   = {Artificial Intelligence and Law: An Overview},
  journal = {Georgia State University Law Review},
  year    = {2019}
}

@article{lewis2020rag,
  author  = {Lewis, Patrick and others},
  title   = {Retrieval-Augmented Generation for Knowledge-Intensive NLP Tasks},
  journal = {NeurIPS},
  year    = {2020}
}

@misc{mastra2024,
  author = {{Mastra AI}},
  title  = {Mastra Framework Documentation},
  year   = {2024}
}

@misc{anthropic2024,
  author = {{Anthropic}},
  title  = {Claude Model Documentation},
  year   = {2024}
}

@misc{katz2023bar,
  author = {Katz, Daniel Martin and others},
  title  = {GPT-4 Passes the Bar Exam},
  year   = {2023}
}

@article{bommarito2017,
  author  = {Bommarito, Michael J. and Katz, Daniel Martin},
  title   = {A Mathematical Approach to the Study of the United States Code},
  journal = {Artificial Intelligence and Law},
  year    = {2017}
}

@article{vaswani2017attention,
  author  = {Vaswani, Ashish and others},
  title   = {Attention Is All You Need},
  journal = {NeurIPS},
  year    = {2017}
}
\end{document}